\documentclass[journal]{IEEEtran}

\usepackage{algpseudocode}
\usepackage{algorithm}
\usepackage{framed,multirow}
\usepackage{epsfig}
\usepackage{subcaption}

\usepackage{changepage}

\usepackage{amssymb}
\usepackage{latexsym}
\usepackage{listings}
\usepackage{graphicx}
\usepackage{overcite}
\usepackage{amsmath}
\usepackage[numbers]{natbib}

 \usepackage{url}
\usepackage{hyperref}
\usepackage{xcolor}

\newcommand{\bfx}{{\bf x}}
\newcommand{\bfy}{{\bf y}}
\newcommand{\bff}{{\bf f}}

\newcommand{\bfv}{{\bf v}}
\newcommand{\bfu}{{\bf u}}

\newcommand{\bfX}{{\bf X}}

\newcommand{\bfY}{{\bf Y}}

\newcommand{\calD}{{\cal D}}
\newcommand{\calT}{{\cal T}}
\newcommand{\calY}{{\cal Y}}
\newcommand{\calX}{{\cal X}}

\def\Re{{\rm I\!R}}                            % the real numbers

\newcommand{\fixme}[1]{}
\newcommand{\update}[2]{#2}

\begin{document}

\title{Domain adaptation for person re-identification on new unlabeled data using AlignedReID++}

\author{Tiago de C. G. Pereira, Teofilo E. de Campos% <-this % stops a space
\thanks{T. Pereira and Teo de Campos are with Departamento de Ci\^encia da Computa\c{c}\~ao, Universidade de Bras\'{\i}lia - UnB, Bras\'{\i}lia-DF, Brazil (e-mail: pereira.tiago@aluno.unb.br; t.decampos@oxfordalumni.org).}}% <-this % stops a space

\maketitle

\begin{abstract}
In the world where big data reigns and there is plenty of hardware prepared to gather a huge amount of non structured data, data acquisition is no longer a problem.
  Surveillance cameras are ubiquitous and they capture huge numbers of people walking across different scenes. 
  However, extracting value from this data is challenging, specially for tasks that involve human images, such as face recognition and person re-identification.
  Annotation of this kind of data is a challenging and expensive task.
  In this work we propose a domain adaptation workflow to allow CNNs that were trained in one domain to be applied to another domain without the need for new annotation of the target data.
  Our method uses AlignedReID++ as the baseline, trained using a Triplet loss with batch hard. Domain adaptation is done by using pseudo-labels generated using an unsupervised learning strategy. 
  Our results show that domain adaptation techniques really improve the performance of the CNN when applied in the target domain. \footnote{This work is an extension from the award winning VISAPP 2020 paper “Domain  adaptation  for  person  re-identification on new unlabeled data" \cite{reid_visapp_2020}}
\end{abstract}

\begin{IEEEkeywords}
Domain Adaptation; Person Re-identification; Deep Learning.
\end{IEEEkeywords}

\section{Introduction}
\label{sec:introduction}

 The purpose of person re-identification is to match images of persons in non-overlapping cameras views. It can be helpful in some important applications as intelligent video surveillance \cite{WANG20133}, action recognition \cite{surveillance} and person retrieval \cite{SVDNet}. 

For problems related to identifying people in images, the first method of choice is usually based on face recognition.
This is because such algorithms have already matched the human capacity, as we can see in Taigman et al.'s work \cite{DeepFace},
where a $97.35\%$ accuracy was achieved in the LFW dataset \cite{LFW} while the human accuracy on the same data is $97.53\%$.
However, face recognition algorithms have little value on surveillance images because the subjects are usually
far away from the cameras, so there is not enough resolution in the area of the face. Furthermore, the surveillance viewpoint
is usually such that a high amount of (self-)occlusion happens, to the point that the faces are not visible at all.
For these reasons, person re-identification algorithms usually take the whole body into account.
The typical workflow to train a person re-identification system follows this steps:
\begin{enumerate}
    \item Use a CCTV system to gather non structured data;
    \item \label{item:filter} Filter this data using a person detector and tracker;
    \item \label{item:annotate} Annotate person bounding boxes;
    \item \label{item:train} Train a metric learning CNN in the annotated data;
    \item Deploy the trained CNN to match people that appear in different cameras.
\end{enumerate}

The biggest problem with this workflow is step~\ref{item:annotate}, because CNNs need a huge amount of data to be properly trained and the process of annotating all the data is very expensive (in terms of time and manpower). We therefore propose to replace this step by an unsupervised domain adaptation technique.
According to Pan and Yang \cite{PAN_YANG_TL}, domain adaptation is a type of transfer learning where only source domain data is labeled and both domains have the same task. 

In our technique, we use a public dataset as our source domain and the non structured data from the CCTV as our target domain. In our source domain all the annotation and image filtering have already been done, then we use unsupervised image-image translation to create an intermediate dataset, the domain-adapted (DA) dataset.
This dataset has the labels of the source domain, but the appearance of people is similar to those in the target domain.
Next, we proceed to the metric learning step using the DA dataset.
As the DA dataset is similar to the target domain, we expect that the CNN trained on it will perform well in the target domain.

In addition, we use this learned metric to annotate the target domain using a clustering algorithm. That way, we have pseudo-labels available for the target domain, then we fine-tune our CNN in these pseudo-labels and learn specific characteristics of the target domain. As the training is performed with the actual target domain images we expect to increase the performance, even though the adaptation process generates a noisy label space for the target domain.

%In our experiments, we evaluated the CNN performance in the target domain (direct transfer), we evaluated the same CNN trained with the dataset adapted by a CycleGAN to the target distribution and we also evaluated the same CNN trained in the target domain using our pseudo label method.
%Our method surpasses the baseline accuracy in all test cases.
%All of that is achieved by replacing step \ref{item:annotate} by our technique and will be explained in more details in sections \ref{sec:proposed_technique} and \ref{sec:results}.

This work is an extension of our article \cite{reid_visapp_2020} presented at VISAPP 2020, where we proved our method's effectiveness. 
In ou previous work, we used a simple and generic network as a feature extractor so we could focus our analysis on the domain adaptation.
In the present work, we replaced the feature extractor by the AlignedReID++ technique proposed by Luo et al.\ \cite{luo2019alignedreid++}. As expected, the use of AlignedReID++ gives much better baseline (direct transfer) results, but we show that our conclusions still hold, i.e., our domain adaptation framework still improves results over AlignedReID++.

%In addition, we observed that the highly unbalanced nature of the person re-identification problem means that training batches may be heavily biased towards negative samples.
%To deal with that, we use a batch scheduler algorithm that allows to train a CNN with a triplet loss in cases where the data is noisy.

In addition, we observed that the highly unbalanced nature of the person re-identification problem means that training batches may be heavily biased towards negative samples.
Previously, we presented a batch scheduler algorithm to deal with that. In the present paper, we dive deeper in this algorithm to understand its real contribution for training a CNN with triplet loss.

Next section discusses related work. Section~\ref{sec:proposed_technique} presents our method and Section~\ref{sec:results} presents experiments and results. This paper concludes in Section~\ref{sec:conclusion}.

\section{Related work}
\label{sec:related}

 The state-of-art in person re-identification follows a pattern of using either attention-based neural networks \cite{Liu_2017_ICCV}, factorization neural networks \cite{Chang_2018_CVPR} or body parts detection \cite{Zhao_2017_CVPR}. The common point in these works is trying to disregard the background information, so they can give the proper weight on the image areas where the person is visible. These methods achieve great results, but have a high complexity, as they are based on combinations of several elements.
In this work we use AlignedReID++ \cite{luo2019alignedreid++}, which uses Dynamically Matching Local Information (DMLI) to align local body parts 
without extra supervision. This eases the CNN job of disregarding background information without increasing complexity.
%However, different datasets have different characteristics and certain combination of methods may not work across all datasets.
%In this paper, our focus is on the exploitation of domain adaptation for this application.
%To design more controlled experiments, we use a relatively simple end-to-end system based on the ResNet-50 \cite{resnet50} as a backbone.

Typically, the person re-identification challenge is approached as a metric learning task \cite{Zhao_2017_CVPR,Deng_2018_CVPR}. But it can also be approached as a classification task where each person from the dataset is a class \cite{Liu_2017_ICCV,Chang_2018_CVPR}.
The problem of the classification-based approach is that the space of labels is fixed and has a large cardinality. Such methods are rarely applicable in practice, unless the set of identities of people who transit through a set of environments is always the same.
Our target application is public spaces, therefore it is not possible to restrict the set of labels. Therefore we approach this as a metric learning challenge\footnote{An alternative would be to model re-identification as a classification problem and use a one-shot learning approach, so that the system can adapt to new people entering the capture spaces. However, this requires that the system first detects if the person is unknown and, in that case, a learning process would have to be triggered at the application stage.}.
Further to being applicable to public spaces, the task of comparing samples is the same across different domains. This  enables the application of unsupervised domain adaptation methods to adapt the marginal distribution of the data.

Recently, some works presented domain adaptations techniques for person re-identification. Zhao et al.\ \cite{Zhao_2017_CVPR} created a new dataset to evaluate the generalization capacity of his model. Their CNN was evaluated in it without further training. Zhong et al.\ \cite{Zhong_2018_CVPR} used a CycleGAN to approximate the camera views in a dataset trying to learn a camera latent space metric. Xiao et al.\ \cite{Xiao_2016_CVPR} trained their CNN with a super dataset created concatenating multiple datasets. They proposed a domain guided dropout to further specialize their CNN for each dataset. In this work, we consider that the target domains have no labeled data, then we cannot use the approaches of Zhong et al.\ \cite{Zhong_2018_CVPR} or Xiao et al.\ \cite{Xiao_2016_CVPR}.
The approach of Zhao et al.\ \cite{Zhao_2017_CVPR} can be called direct transfer, because it just evaluates a CNN on a target domain.
We shall demonstrate that our method outperforms direct transfer.

\section{Proposed Method}
\label{sec:proposed_technique}

 Our technique is based on training a CNN to learn a metric, so we can ensure that distinct domains will have the same task. Therefore, we train 
AlignedReID++ (Section \ref{subsec:CNN}) with the triplet loss (Section \ref{subsec:triplet}) to learn the desired metric in a Euclidean vector space. 
Also, we evaluate the contribution of a batch scheduler algorithm to deal with noisy datasets (Section \ref{subsec:batchScheduler}).
The core of the domain adaptation method is based in a CycleGAN that will perform an image-image translation to approximate source and target domains 
(Section \ref{subsec:cycleGAN}). Then, we use the CNN trained in the domain-adapted dataset to extract the features of the target domain images and use an unsupervised learning algorithm to generate pseudo-labels for the target domain (Section \ref{subsec:kmeans}).

\subsection{AlignedReID++}
\label{subsec:CNN}

As said in Section \ref{sec:related}, the state-of-art in person re-identification use techniques that exploit information from CNNs at multiple levels, 
bringing multiple semantic levels to the final features. Those semantic levels may carry specific person attributes like gender, textures and clothing, 
which are important for matching people across views.

The AlignedReID++ \cite{luo2019alignedreid++} uses Resnet-50 as a feature extractor and propagates its output to two branches, local and global.
The final convolutional layer produces a feature map with dimensions $C \times H \times W$ ($C$ is the number of channels
and $H \times W$ is the spatial size). This feature map is the information that is propagated to both branches.

For the global branch, a global average pooling is used to reduce the feature map into a global feature vector with size $C \times 1$. Then, this 
global feature vector is used to calculate a Softmax Loss ($\mathcal{L}_{ID}$) and to calculate the global distances that will be used by the global 
triplet loss ($\mathcal{L}^{g}_{T}$).

\begin{figure}[htpb]
    \centering
    {\includegraphics[width=1.0\columnwidth]{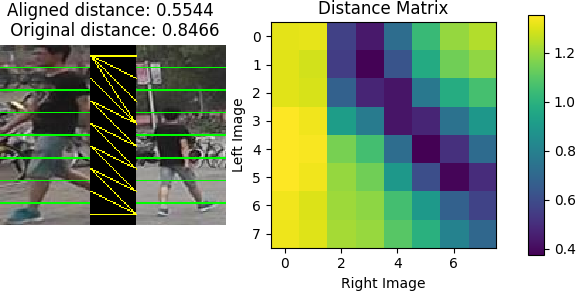}}
    \caption{Example of how AlignedReID++'s Dynamically Matching Local Information (DMLI) is able to align two pictures that were displaced because of the camera views.
      The distance matrix on the right is computed by comparing stripes of the two images and their minimum path on that matrix generates the alignment shown on the left. As expected, the aligned distance is smaller than the global distance.
            The code used to generate this image is available from \url{https://github.com/michuanhaohao/AlignedReID}}
    \label{fig:aligned}
\end{figure}

The local branch uses a horizontal max pooling to reduce the feature map into a $C \times H \times 1$ local feature map, which is further reshaped into 
the size of $H \times C$. The local feature maps are then split into horizontal regions (strips) and compared with all the horizontal strips from other image to calculate 
a distance matrix. This distance matrix has the size $H \times H$ and is used to calculate the shortest path from $(1 \times 1)$ to $(H \times H)$.
This method is called Dynamically Matching Local Information (DMLI) and provides a local distance (shortest path) between two local feature maps. The local distances 
are then used to calculate the local triplet loss ($\mathcal{L}^{l}_{T}$).

The local branch is able to align parts of the image that may be displaced because of the camera view (Fig.~\ref{fig:aligned}). The global branch
is able to extract the global image context and a class biased information (softmax loss). Finally, the AlignedReID++ loss is a combination of these 3 losses given by
Eq.~\ref{eq:aligned}.
\begin{equation}
    \mathcal{L} = \mathcal{L}_{ID} + \mathcal{L}^{l}_{T} + \mathcal{L}^{g}_{T}
    \label{eq:aligned}
\end{equation}

%, because we believe that residuals blocks help to propagate information from multiple semantic levels when they are relevant for the output. Although the residual blocks may not perform as well as a specific architecture, the main goal of our work is to propose a domain adaptation workflow.

%To have an \update{\em jump start}{initial boost} \cite{DeCAF}, we start with a ResNet-50 CNN pre-trained on ImageNet \cite{imagenet_cvpr09}.
%We then transfer learn it to the problem of person re-identification using a public dataset. This is done by replacing 
%the last fully connect layer by a new fully connected layer with 128 features which are used as an embedding for
%metric learning. We use Adam optimizer and the triplet loss.

\subsection{Triplet Loss and Batch Hard}
\label{subsec:triplet}

A siamese-like loss is ideal when trying to learn a metric because it allows one to perform an end-to-end learning from a dataset to an embedding space. The siamese loss receives as input a pair of feature vectors and tries to approximate them  if they are from the same person or set them apart if they are from different people. 
This generates an embedding space where feature vectors from the same person tend to lie near each other.

The triplet loss is an upgrade from the  siamese loss which instead of using a pair of samples as input, it uses an anchor, a positive sample and a negative sample. Therefore, the triplet loss approximates feature vectors from the same person while it also separates features of different people, according to Equation \ref{eq:triplet} (defined for each anchor sample $\bfx_a$). This way, one can expect better samples separation in the embedding space:
\begin{equation}
    \label{eq:triplet}
    \mathcal{L}(\bfx_a) = \max \bigg( 0 ~, ~ m + D \Big( \bff_a, \bff_p \Big) - D \Big( \bff_a, \bff_n \Big) \bigg),
\end{equation}
 where $m$ is a margin so the loss does not go to zero, $\bff$ is the CNN output, i.e., a lower dimensional embedding of image $\bfx$; (sub indexes $a$, $p$ and $n$ mean anchor, positive and negative, respectively) and $D(\cdot)$ can be any distance measurement algorithm, in our case is the Euclidean distance defined by
\begin{equation}
    D(\bfu, \bfv) = \sqrt{\sum_{i=1}^{d}{(u_i - v_i)^2}}.
    \label{eq:euclidiana}
\end{equation}

A question that arises from the triplet loss use is ``how to choose the positive/negative examples?''
Hermans et al.\cite{defense} investigated this problem and came to a conclusion that the best learning is achieved when using the hardest positive/negative samples during training.
This approach was coined {\em batch hard} and it works as follows: for each anchor sample $\bfx_a$ from the batch, the choice of positive sample $\bfx_p$ is chosen as the one that maximizes $D(\bff_a, \bff_p)$ and the negative sample $\bfx_n$ is chosen as the one that minimizes $D(\bff_a, \bff_n)$. Using this strategy, Equation \ref{eq:triplet} can be rewritten as
\begin{align}\label{eq:triplet_bh}
    \mathcal{L}_{BH}(\bfx_a) = \max \bigg( 0 ~ , ~ m 
    & + \max_{p} {D \Big( \bff_a, \bff_p \Big)} 
    \\
    & - \min_{n} {D \Big( \bff_a, \bff_n \Big) \bigg)} , \nonumber
\end{align}
where positive and negative samples are chosen within each batch and the losses across all anchors in a batch are averaged out.

\begin{figure}[htpb]
    \centering
    {\includegraphics[width=.9\columnwidth]{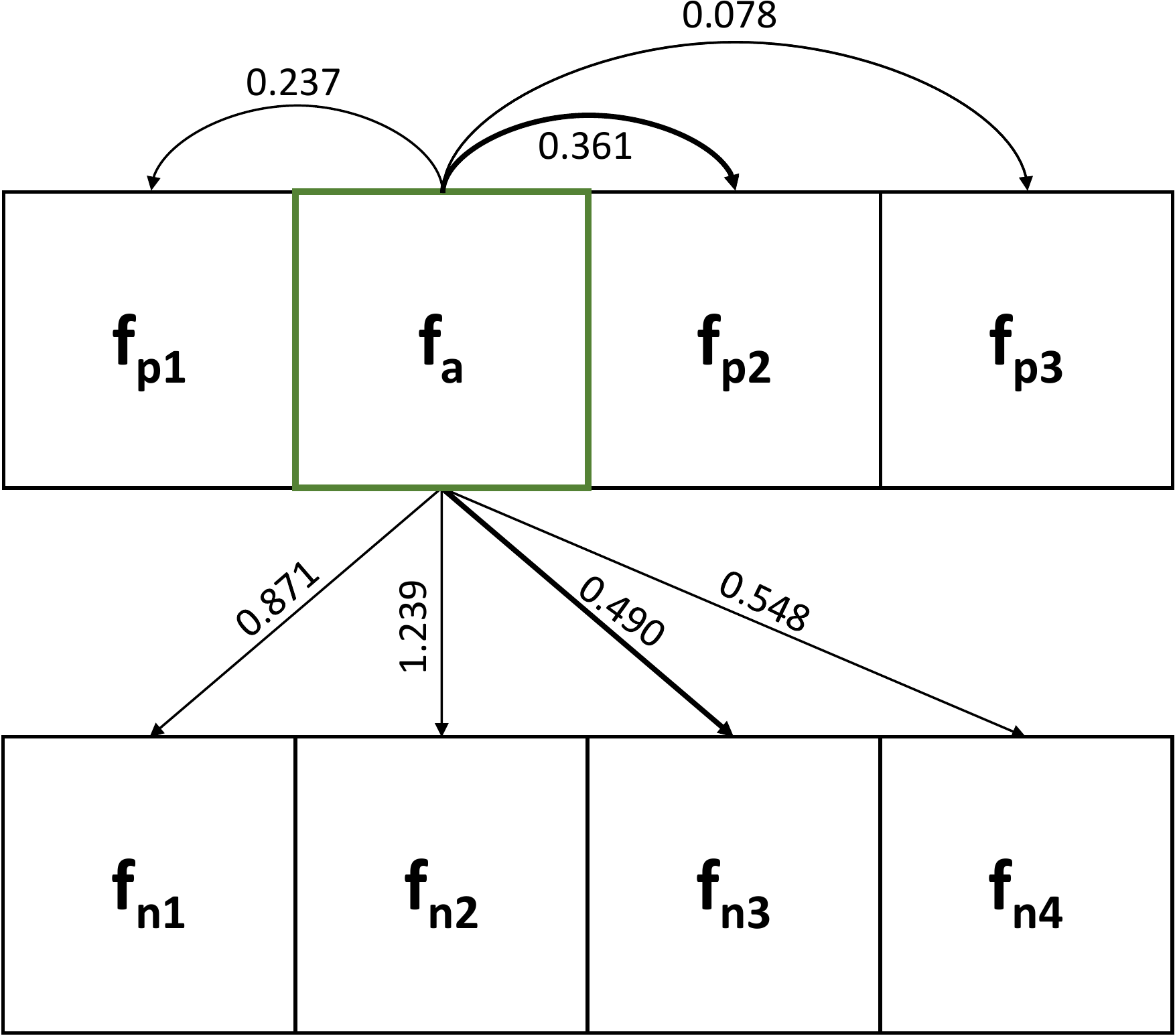}}
    \caption{Example of a batch hard triplet selection.}
    \label{fig:ex_bh}
\end{figure}

Figure \ref{fig:ex_bh} illustrates how samples are chosen for a batch.
All the rectangles at the top represent samples from a person and the rectangles at the bottom represent sample of another person.
The triplet will choose each rectangle as anchor at a time, calculate the loss for it and in the final sum all the losses.
From the green rectangle as an anchor, the numbered arrows indicate the distance $D(\cdot)$ from it to the samples,
where $\bff_{p_i}$, $i = \{1, 2, 3\}$, are possible positive samples and $\bff_{n_j}$, $j = \{1, 2, 3, 4\}$, are the possible negative samples.
In a batch hard approach, $\bff_{p_2}$ is selected as positive sample, $\bff_{n_3}$ as negative sample and $\mathcal{L}_{Tri} = m + 0.361 - 0.490$.
% From the green rectangle as an anchor, the numbered arrows indicate the distance $D(\cdot)$ from it to the samples,
% where $\text{Pos\_i, i} = 1, 2, 3$, are possible positive samples and $\text{Neg\_j, j} = 1, 2, 3, 4$, are the possible negative samples.
% In a batch hard approach, Pos\_2 is selected as positive sample, Neg\_3 as negative sample and $\mathcal{L}_{BH} = m + 0.361 - 0.490$.

\subsection{Batch Scheduler}
\label{subsec:batchScheduler}

Hermans et al.\cite{defense} proved the batch hard effectiveness, but choosing the
hardest samples at each batch increases the training complexity.
Furthermore, we work with a domain adapted dataset that can be noisy,
meaning that the separation between positive and negative samples may
be less trivial, which increases the training cost even more.
The consequence is that the training process may never converge with this strategy.
When using the triplet loss, a non converging training process can be
identified if the loss is stuck at the margin ($m$),
because that means $D ( \bff_a, \bff_p) = D ( \bff_a, \bff_n)$,
meaning that all the features are converging to vectors of 0s.

While training with the triplet loss, the goal is to make $D ( \bff_a, \bff_p) < D ( \bff_a, \bff_n)$.
However, if the batch is big, the number of negative examples is way bigger than the number of positive examples,
particularly in the case of person re-identification.
It is therefore possible to have a negative sample that is nearer to the anchor than the hardest positive sample.
This way the loss will always be greater than the margin ($\mathcal{L}_{BH} > m$), then the optimizer learns that
outputting vectors  of 0s will reduce the loss to the margin, i.e., ($\mathcal{L}_{BH} = m$).

Our solution was to use a batch scheduler algorithm to decrease the number of negative samples and lower the training complexity.
This way we ease the training convergence, and once the training is converging we slowly increase the batch size (and therefore its complexity, having an impact in the loss).
This enables us to learn step by step and converge the training even with a noisy dataset.
Our batch scheduler algorithm is shown in Algorithm \ref{alg:bs}.

\begin{algorithm}
    \caption{Batch Scheduler}
    \label{alg:bs}
  
    \begin{algorithmic}[1]
      \State $batch\_size = 2 \times n\_instances$
      \State $m = 0.3$ %~ ~ ~ // $m$ is the loss margin of Eq.~\ref{eq:triplet}
      \For  {$i=0$ to $num\_epochs$} 
      \State        $loss = train(i, batch\_size)$ 
                    \If{$loss < (0.8 \times m)$}
      \State            $batch\_size = batch\_size \times 2$
                    \EndIf
      \EndFor
    \end{algorithmic}
 \end{algorithm}

In Algorithm \ref{alg:bs}, $m$ is the loss margin of Eq.~\ref{eq:triplet} and $n\_instances$ is the number of samples 
for each person ID, we used $n\_instances = 4$. The training start with samples from 2 person IDs per batch. When
$\mathcal{L}_{BH} < m$ the training converged, because this is only possible 
if the CNN can distinguish the person IDs, as shown in 
Eq.\ \ref{eq:condition}. In line 5 of the algorithm we used a $0.8$ factor to ensure this convergence.

\begin{equation}
    \mathcal{L}_{BH} < m \Leftrightarrow D ( \bff_a, \bff_p) < D ( \bff_a, \bff_n)
    \label{eq:condition}
\end{equation}

Once the convergence is ensured, we can go one step further and increase the training complexity. Then, we double the 
batch size, doubling the number of person IDs per batch. This process is repeated until we reach the final epoch or 
the maximum GPU memory. 

For this work, we used a NVDIA GTX 1070 Ti GPU with 8 GB of VRAM, so the maximum batch we could 
reach had 88 images (22 person IDs). We recognize this still is a small batch and recommend experiments to use up to 
256 images per batch.

Smith et al. \cite{DONT_DECAY} argue that increasing the batch size instead of decreasing the learning rate results in a 
faster training convergence. This argument is based in the scale of random fluctuations in the optimizer given by Eq.~\ref{eq:noise}.
Where $N$ is the training set size, $B$ represent the batch size and $\varepsilon$ is the learning rate.

\begin{equation}
    g = \varepsilon \bigg( \frac{N}{B} - 1 \bigg)
    \label{eq:noise}
\end{equation}

Assuming a big training set $N$, Eq.~\ref{eq:noise} can be approximated by $g \approx \varepsilon N/B$. Therefore,
increasing the batch size or decreasing the learning rate should have the same impact in the noise scale. However, increasing the 
batch size leads to a significantly reduction in the number of parameter updates needed, speeding up the training. 

Also, the initial high noise scale allows us to explore a larger fraction of the loss function without becoming trapped in local 
minima. This way, we believe that the slow increase in the training complexity may lead us to a better region in the parameter space.
Therefore, we reduce the noise scale and fine-tune the parameters to find the promising local minima.

\subsection{Image-Image Translation for Domain Adaptation}
\label{subsec:cycleGAN}

To give some background, the definitions and notations used in this paper
are based on \cite{csurka_comprehensive_2017} and \cite{PAN_YANG_TL}. 
A domain $\calD$ is composed of a $d$ dimensional feature space $\calX \subset \Re^d$  with a marginal probability distribution $P(\bfX)$ and a task $\calT$  defined  by a label space $\calY$ and the conditional probability distribution $P(\bfY|\bfX)$, where $\bfX$ and $\bfY$ are sets of random variables (which usually are multivariate). Given a particular sample set  $\bfX=\{\bfx_1,\cdots, \bfx_n\} \in \calX$, with corresponding labels $\bfY=\{\bfy_1,\cdots, \bfy_n\} \in \calY$, $P(\bfY|\bfX)$ in general can be learned in a supervised manner from these feature-label pairs  $\{\bfx_i, \bfy_i\}$.

For simplicity, let us assume that there are two domains: a source  domain $\calD^s=\{\calX^s,P(\bfX^s)\}$ with $\calT^s=\{\calY^s,P(\bfY^s|\bfX^s)\}$ and a target domain  $\calD^t=\{\calX^t,P(\bfX^t)\}$ with $\calT^t=\{\calY^t,P(\bfY^t|\bfX^t)\}$. Those domains are different $\calD^s \neq \calD^t$, because $P(\bfX^s) \neq P(\bfX^t)$ due to domain shift. Also, we do not have the target domain labels $\bfY^t$, so we do not have the feature-label pairs  $\{\bfx_i, y_i\}$ to learn $P(\bfY|\bfX^t)$ in a supervised manner.

The person re-identification task $\calT$ consists in learning a projection from $\bfx\in\calX$ to a feature $\bff$ in a Euclidean space where $\bff$ is closer to other vectors if they originated from the same person, more distant to vectors from other people.
The set of labels can be thought of as the space of all possible person identities in the world, which impractical. Alternatively, the person re-ID problem can be seen as a binary problem that takes two samples as input, indicating whether or not they come from the same person.
Therefore, each person re-ID dataset (or indeed each camera surveillance environment) can be seen as a different domain, however the task is always the same, i.e., telling if two images contain the same person or not.
Domain adaptation are transductive transfer learning methods where it is assumed $\calT^s = \calT^t$, according to Csurka \cite{csurka_comprehensive_2017}. Therefore, we can use domain adaptation to exploit the related information from $\{\calD^s,\calT^s\}$ to learn  $P(\bfY^t|\bfX^t)$.

In our method, we have images from source domain $\bfX^s$ and target domain $\bfX^t$, but we do not have the labels from target domain $\calY^t$. So, we approximate data from images of a known source domain to images of a target domain generating an intermediate (DA) dataset. 

We use, as source domain, a public dataset which has ground truth annotation of positive/negative examples for each anchor.
An unsupervised domain adaptation method can be used to generate an intermediate dataset $\calD^i$ that leverages the source domain annotation $\calY^s$ and is similar to the target domain.
For that, we follow an approach based on Generative Adversarial Networks -- GANs \cite{GAN_Goodfellow}.
\update{More specifically, we use a CycleGAN using the method proposed by Zhu et al.\ \cite{Unpaired_ImageToImage_CycleGAN}.}{More specifically, we use the CycleGAN method proposed by Zhu et al.\ \cite{Unpaired_ImageToImage_CycleGAN} and applied to person re-identification by Deng et al.\ \cite{Deng_2018_CVPR}.}

The idea is to use images from the source domain ($\bfX^s$) as input and train a GAN to generate outputs which are similar to the images from the target domain ($\bfX^t$). However, once we have no paired images between domains the problem has a high complexity. Zhu et al.\ proposed to train two generators $G$ and $F$ where $G: \calX^s \rightarrow \calX^t$ is a mapping from the source domain to the target and $F: \calX^t \rightarrow \calX^s$ is a mapping from the target domain to the source. Also, a cyclic component is added to the loss:
\begin{align}
    \mathcal{L}(G,F,D_{\calX^s}, D_{\calX^t}) = \ & \mathcal{L}_{GAN}(G,D_{\calX^t},\bfX^s,\bfX^t) + \nonumber \\ 
    & \mathcal{L}_{GAN}(F,D_{\calX^s},\bfX^t,\bfX^s) + \\ 
    & \lambda \mathcal{L}_{cyc}(G,F) ,\nonumber
    \label{eq:custo_cycle}
\end{align}
 where both $\mathcal{L}_{GAN}$ components are the basic GAN loss proposed by Goodfellow et al.\ and the $\mathcal{L}_{cyc}$ is the cyclic component added by Zhu et al., wich is given by:
\begin{eqnarray}
    \mathcal{L}_{cyc}(G,F) =  E_{\bfX^s \sim p_{data}(\calX^s)} \big[ \left \| F(G(\bfX^s)) - \bfX^s \right \|_1  \big] + \nonumber\\
                   E_{\bfX^t \sim p_{data}(\calX^t)} \big[ \left \| G(F(\bfX^t)) - \bfX^t \right \|_1  \big] 
    \label{eq:custo_cyc}
\end{eqnarray}
 the cyclic component is there to do an identity match between  source domain images $\bfX^s$ and their double transformed pairing images $F(G(\bfX^s))$, and vice-versa. By minimizing this cyclic loss we expect to have transformations that can map both domains. 

Therefore, we use the generator $G: \calX^s \rightarrow \calX^t$ in all images of our source domain to generate an intermediate dataset. That is, we create a dataset that leverages from the labeled data of the source domain and have similar characteristics to the target domain. This way we can expect that a training on the DA dataset will perform well in the target domain.

\subsection{Pseudo-Labels for Re-Identification}
\label{subsec:kmeans}

In Section \ref{subsec:triplet}, we used the triplet loss to learn a distance metric in an Euclidean vector space. In Section \ref{subsec:cycleGAN}, we showed that both source and target domains have the same label space $\calY$. We also presented a method to train our CNN in an intermediate dataset that leverages from the labeled data of the source domain and have similar characteristics to the target domain. The CNN therefore should already present a reasonable performance in target domain.

We use the CNN to extract all features $\bff_i^t$ from target domain images $\bfX^t$ and these features belong to an Euclidean vector space. Then, we used a clustering algorithm to group these features, using the obtained group identifications as target domain with pseudo-labels $\bfY^t$. 
In addition, we fine tune the CNN using the feature-label pairs  $\{\bfx_i, y_i\}$ with the real images from target domain and the pseudo-labels generated by the clustering algorithm.

Even though the pseudo labels generated may contain a lot of errors, this next training step uses the real images from target domain $\bfX^t$. Therefore, the CNN is be able to learn more robust features for the target domain, because it learns the exact characteristics of the target domain.

We choose the k-means \cite{kmeans} clustering algorithm to group the features in the Euclidean vector space. The value of $k$ was chosen as a proportion of the size of each target dataset. Table \ref{table:kmeans} indicates the values used in this paper (the datasets are discussed later). However, the naive assignment of samples to clusters is a flawed strategy to annotate the data, because a simple look at the data may cluster viewpoints rather than people. 
In other words, features from different people taken from the same camera view are often more similar to each other than features from the same person from different camera views.

\begin{table}[htpb]
    \centering
    \caption{The chosen $k$ for each dataset when using k-means algorithm.}

    \begin{tabular}{|c|c|}
    \hline
    \textbf{Dataset} & \textit{k} \\ \hline
    CUHK03                   & 2000       \\ \hline
    Market1501               & 1600       \\ \hline
    Viper                    & 632        \\ \hline
    \end{tabular}
    \label{table:kmeans}
\end{table}

Our solution is to use k-means algorithm to generate k clusters for each camera view, then use a nearest neighbor algorithm to group these clusters across the camera views. This way, we guarantee that every person from our pseudo-labels space have images from each camera. That results in a noisy annotation, because that assumption is not a true in the real label space of the dataset. However, using this approach we ease the CNN task of learning features robust for multiple cameras views and achieve better results in validation.

\section{Experimental results}
\label{sec:results}

In our work, we produced results using three well known person re-identification datasets: CUHK03 \cite{deepReID}, Market1501 \cite{zheng2015scalable} and Viper \cite{viper}. For all the experiments, we did not use any label information in the target domain, except to  evaluate the results. 

Figure~\ref{fig:GAN_montage} shows qualitiative results of domain adaptation, discussed in Section~\ref{subsec:cycleGAN_results}.
All quantitative results are shown in Table~\ref{table:results}, which are discussed in the subsequent sections.
% Sections~\ref{subsec:result_batch} and \ref{subsec:domain_results}.

\subsection{Qualitative results on the Domain-Adapted dataset}
\label{subsec:cycleGAN_results}

As said in Section \ref{subsec:cycleGAN} our method tries to approximate the source domain to the target domain. This is done training a CycleGAN between both domains and using the generator to create an intermediate dataset that shifts the source domain samples so that they become more similar to the target domain data. The idea is to generate images that preserve the person morphology, but are visually adapted to the target domain.
While there is no guarantee that a GAN preserves person morphology, the cyclic loss contributes towards this goal, as it has an identity match component.

\begin{figure*}[htpb]
    \centering
    {\includegraphics[width=1\textwidth]{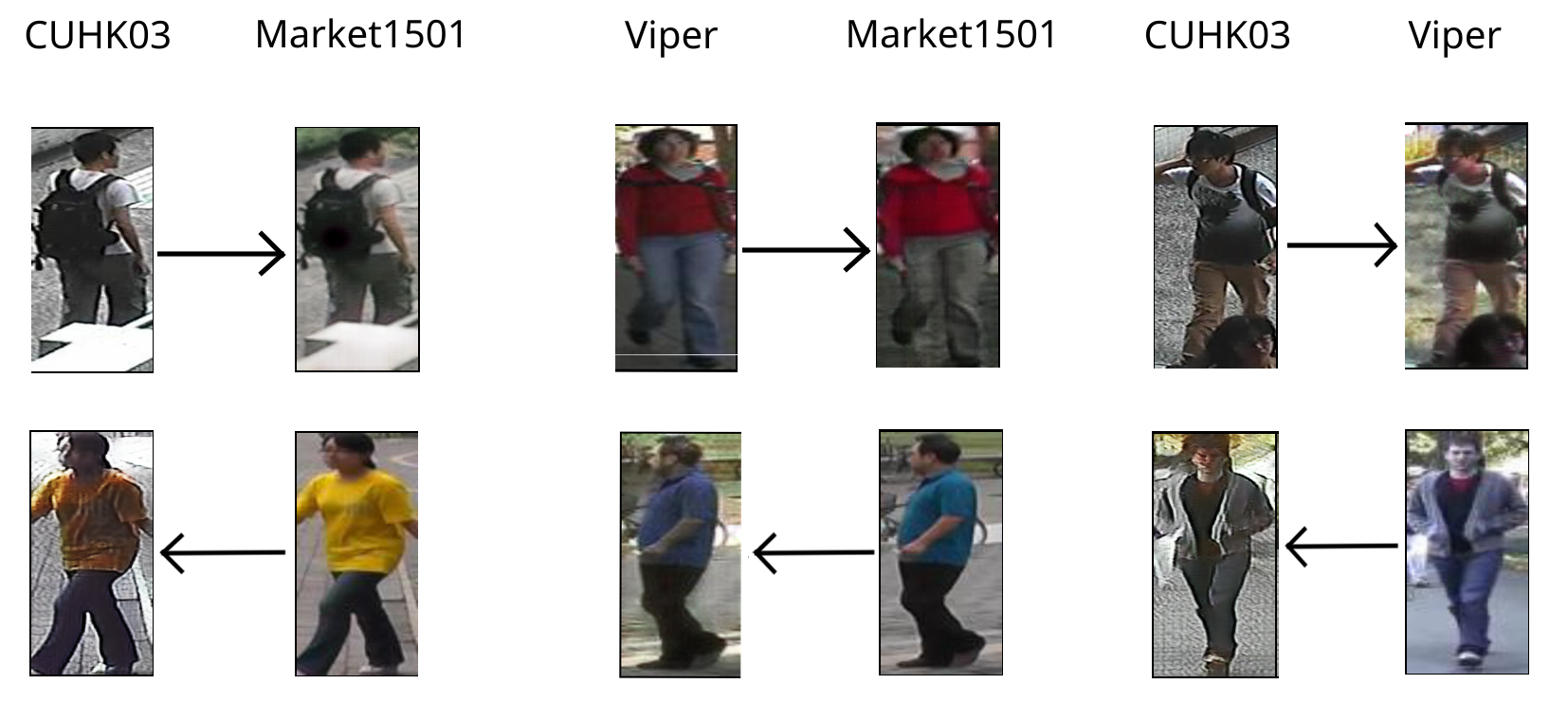}}
    \caption{Examples illustrating domain adaptation between domains using CycleGAN.}
    \label{fig:GAN_montage}
\end{figure*}

Figure \ref{fig:GAN_montage} presents examples of transformation results between all domains. It is interesting to note that the morphology of people has been preserved in all images and the changes have more effect in the colors, texture and background. That means we could produce a great approximation of how a person would look like in the view of another dataset.

The CUHK03 dataset was created using surveillance cameras from a university in Hong Kong with an elevated viewpoint, so normally the background of their images consists in a granular floor. The Market1501 dataset was created with cameras in a park, so the images usually have grass in the background of their views. Viper is the oldest dataset used in this work, it was published in 2007 and is composed of low resolution outdoor images. 

These characteristics of the datasets make it easy to understand the effects seen in Figure \ref{fig:GAN_montage}.  When using CUHK03 as the target domain, the transformed images tend to have a granular background to approximate the floor texture in CUHK03 images. When using Market1501 as target domain, images from CUHK03 had a background transformation from the granular floor to grass, and images from Viper had just a color transformation, because both datasets are from outdoor images. When using Viper as target domain, images from Market1501 had a color transformation and images from CUHK03 had a texture background transformation and a brightness enhancement.

%For Viper as target, I used the CUHK03 metric
\begin{table*}[htpb]
    \centering
    \caption{\label{table:results}CMC accuracy results (in \%) using Rank-1, Rank-5 and Rank-10, obtained using one dataset as source domain and another as target.
      The Work column indicates if the results came from our previous paper\cite{reid_visapp_2020} (which uses a basic ResNet as backbone) or from the present paper (New -- which uses AlignedReID++). Cells containing two values show results without/with our batch scheduler.
    As for the methods, Direct refers to application without transfer and Ours is the combination of CycleGAN and pseudo-labels for domain adaptation.}
    %\begin{adjustwidth}{-1cm}{}
    % \resizebox{\textwidth}{!}{%
        \begin{tabular}{|c|c|c|c|c|c|c|}
    
    \hline
    \multicolumn{4}{|l|}{}                                                   & \multicolumn{3}{c|}{\textbf{CMC Accuracy} (in \%)}               \\ \hline
    \textbf{Source}               & \textbf{Target}   & \textbf{Work}            & \textbf{Method}          & \textbf{Rank-1} & \textbf{Rank-5} & \textbf{Rank-10} \\ \hline
    \multirow{12}{*}{\textbf{Market1501}} & \multirow{6}{*}{\textbf{Viper}} & 
    
    %% MARKET 1501  --->  Viper
    \multirow{3}{*}{\textbf{Previous\cite{reid_visapp_2020}}}       & Direct            & 12.5      &       25.0         &           33.1      \\ \cline{4-7} 
    &                               &                           & CycleGAN          & 9.8     &       26.9       &           36.4      \\ \cline{4-7}
                                       &          &             & Ours              &  13.9   &       29.0       &            40.7      \\ \cline{3-7}
                                       &  & 
    \multirow{3}{*}{\textbf{New}}                            & Direct            & 22.9      & 41.8               &        50.0         \\ \cline{4-7} 
                         &                                    & & CycleGAN          & 21.4 / 22.8  & 40.2 / 39.1           &     50.3 / 48.9             \\ \cline{4-7}
                      &                                  &      & \textbf{Ours}     & \textbf{23.7} / 21.5  &       41.5 / \textbf{41.9}     &           50.8 / \textbf{51.3}       \\ \cline{2-7}
                                      & \multirow{6}{*}{\textbf{CUHK03}} &   
    
    %% MARKET 1501  --->  CUHK03                           
    \multirow{3}{*}{\textbf{Previous\cite{reid_visapp_2020}}}       & Direct            &  19.9     &              49.4  &              63.2   \\ \cline{4-7} 
                                    &                   &       & CycleGAN          &  34.8     &      66.7          &          79.1       \\ \cline{4-7}
                                    &                   &       & Ours     & 38.2      &       69.7         &           81.6      \\ \cline{3-7}
                                    &  &   
    \multirow{3}{*}{\textbf{New}}                            & Direct            & 22.5       & 45.0               &        58.0         \\ \cline{4-7} 
                                                            &  && CycleGAN          &  37.0 / 38.9   & 69.1 / 69.2           &       80.9 / 81.1          \\ \cline{4-7}
                                                        &&      & \textbf{Ours}     & 42.9 / \textbf{43.1}           &       72.5 / \textbf{72.7}     &           81.2 / \textbf{84.2}       \\ \hline
    \multirow{12}{*}{\textbf{CUHK03}}   & \multirow{6}{*}{\textbf{Viper}} &  
   
    %%   CUHK03 --> Viper
    \multirow{3}{*}{\textbf{Previous\cite{reid_visapp_2020}}}       & Direct            & 10.1       & 22.5               &    29.0            \\ \cline{4-7} 
    &                   &                                       & CycleGAN          & 11.6       & 25.5               &       34.7         \\ \cline{4-7}
                                        &                     & & Ours     & 13.6   & 33.9               &           46.0      \\ \cline{3-7}
                                        &  &  
    \multirow{3}{*}{\textbf{New}}                            & Direct            & 20.6      & 38.0               &          47.2       \\ \cline{4-7} 
                                          &                &    & CycleGAN          & 21.8 / 17.9  & 43.2 / 39.9           & 52.2 /  50.9               \\ \cline{4-7}
                                         &                    & & \textbf{Ours}     & \textbf{22.5} / 18.5  &       \textbf{43.2} / 38.0     &           \textbf{54.1} / 50.2       \\ \cline{2-7}
                                                               & \multirow{6}{*}{\textbf{Market1501}} & 
   
    %%   CUHK03 --> Market1501                                                          
   \multirow{3}{*}{\textbf{Previous\cite{reid_visapp_2020}}}        & Direct            & 26.8      & 45.9               &        55.1         \\ \cline{4-7} 
                                                          &  &  & CycleGAN          & 35.8      & 56.5               &        65.7         \\ \cline{4-7}
                                        &                     & & Ours     & 37.3      & 60.4               &        70.4      \\ \cline{3-7}
                                        &  & 
   \multirow{3}{*}{\textbf{New}}                             & Direct            & 38.7      & 55.1               &        62.6         \\ \cline{4-7} 
                                                           &  & & CycleGAN          & 42.7 / 38.4  & 59.7 / 57.2           &       67.3 / 65.5          \\ \cline{4-7}
                                        &                     & & \textbf{Ours}     & 46.8 / \textbf{50.1}  &       65.9 / \textbf{68.2}     &           73.6 / \textbf{75.6}       \\ \hline
    \multirow{12}{*}{\textbf{Viper}} &   \multirow{6}{*}{\textbf{CUHK03}} &  
    
    %% Viper --> CUHK03
    \multirow{3}{*}{\textbf{Previous\cite{reid_visapp_2020}}}       & Direct            & 5.9      & 18.1               &          29.0       \\ \cline{4-7} 
                        &                &                      & CycleGAN          & 31.9     & 64.4               &           77.5              \\ \cline{4-7}
                                         &                    & & \textbf{Ours}     & \textbf{36.1}     & \textbf{69.2}               &           \textbf{81.3}     \\ \cline{3-7}
                                         &   &  
    \multirow{3}{*}{\textbf{New}}                            & Direct            & 9.9      & 27.9               &          40.1       \\ \cline{4-7} 
                                          &                &    & CycleGAN          & 17.1 / 14.5  & 41.6 / 33.5           & 55.8 /  45.7               \\ \cline{4-7}
                                         &                    & & Ours     & 20.4 / 17.5  &       43.9 / 44.5     &           58.5 / 59.5       \\ \cline{2-7}
                                                            & \multirow{6}{*}{\textbf{Market1501}} & 
                                                            
    %% Viper --> Market1501                                                        
    \multirow{3}{*}{\textbf{Previous\cite{reid_visapp_2020}}}       & Direct            & 5.7      & 15.5               &     22.2         \\ \cline{4-7} 
                                                           &  & & CycleGAN          & 6.7      & 17.0              &     23.7            \\ \cline{4-7}
                                         &                 &    & Ours     & 6.6      & 20.5                  &     28.4      \\ \cline{3-7}
                                         &  & 
    \multirow{3}{*}{\textbf{New}}                            & Direct            & 15.9      & 28.2               &        35.4           \\ \cline{4-7} 
                         &                                    & & CycleGAN          & 23.1 / 11.2  & 37.9 / 22.6           &     45.8 / 29.2             \\ \cline{4-7}
                      &                                  &      & \textbf{Ours}     & \textbf{28.4} / 27.6  &       \textbf{46.4} / 43.9     &           \textbf{55.2} / 52.4       \\ \hline
    \end{tabular} % }
    %\end{adjustwidth}
\end{table*}

\subsection{Image-image translation method for domain adaptation}
\label{subsubsec:imageimage}

After successfully generating an intermediate dataset that approximates both domains we use it to fine-tune the CNN trained in the source domain. We evaluated all the results in the target domain using the CMC score with rank-1, rank-5 and rank-10, as shown in the ``CycleGAN'' rows of Table~\ref{table:results}.
That method was compared with the direct transfer method, which consists in evaluating a CNN trained in the source domain directly applied on the target domain, without further training. The direct transfer method therefore shows how different both domains are and is used as a baseline.

As one can see, the CycleGAN method presents huge rank-1 improvements when using CUHK03 as target domain ($7.2\%$ improvement for Viper as source domain and $16.4\%$ improvement for Market1501 as source domain). This happens because the CUHK03 images have granular background texture as a strong characteristic that was easily learnt by our CycleGAN.

Good rank-1 improvements were also obtained for Market1501 as target domain, where the CycleGAN method achieved a $4\%$ improvement for CUHK03 as souce domain and $7.2\%$ for Viper as source domain. These improvements shows that the color transformation helped to approximate these domains, but this was not as significant as texture changes that occurred when working with CUHK03 images.

For Viper as a target domain the CycleGAN method achieved $1.2\%$ rank-1 improvement using CUHK03 as source domain and $0.1\%$ rank-1 decrease for Market1501 as source domain. Again, this means that texture transformations are more significant than color transformations. Those results point to how difficult is the task of creating an intermediate dataset in a unsupervised manner without much data.

Furthermore, when comparing our new results with those of the previous work, the AlignedReID++ contribution is clear.
Using this state-of-art method as a feature extractor allowed us to achieve improvements from $4.1\%$ up to $16.4\%$. The only 
domain combination that did not lead to an improvement was using the Viper as source domain and CUHK03 as target domain.
%Acho que devia argumentar esse problema da VIPER -> CUHK03, mas nao sei pq isso ocorreu..

\subsection{Pseudo-Labels Method}
\label{subsubsec:pseudo}

Section \ref{subsubsec:imageimage} proved the effectiveness of domain adaptation and that the CycleGAN successfully shifted images to the target domain appearance, carrying their source label with them. Also, it was clear that texture transformations are more significant than color transformations.

\begin{figure*}[htpb]
    \centering

    {\includegraphics[width=1\textwidth]{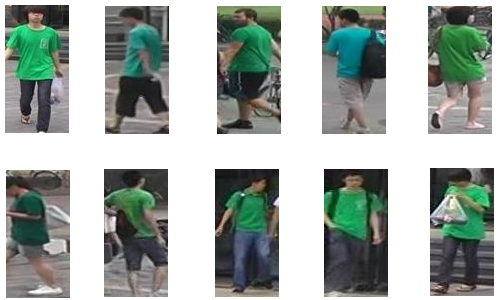}}

    \caption{Images from a final cluster when using the pseudo-labels method. The cluster was achieved using Viper as source dataset and Market1501 as target dataset.}
    \label{fig:pseudo_montage}
\end{figure*}

Although the CycleGAN does a great job at shifting images between domains, when using the pseudo-labels method we achieved even better results. This is because the training is now performed with the actual target domain images and estimated pseudo-labels. So, there is no longer the problem of images in which the person morphology was not preserved. The target dataset characteristics are better represented. Figure~\ref{fig:pseudo_montage} illustrates the dataset created using pseudo-labels -- as one can see the estimated labels are not perfect, but the grouped  images show a strong color similarity.

As one can see in the ``Ours'' rows of Table \ref{table:results}, our pseudo-labels method showed great improvements in all test cases, when compared to the CycleGAN method. Even using the Viper dataset as target domain our method could improve the CycleGAN results in $0.7\%$ or more. For the Market1501 dataset the rank-1 improvement was around $5\%$ to $8\%$ also and for the CUHK03 our method achieved improvements of $3\%$ to $4\%$ in rank-1 accuracy.

%It is important to notice that the pseudo-labels have a stronger positive impact on smaller target datasets. This is because small datasets require fewer clusters to annotate the data. This was very significant for the great results presented for Viper dataset.

It is important to notice the great contribution that AlignedReID++ bring to our method, with improvements up to $19.8\%$. Although the result with Viper as source domain and CUHK03 
was not the expected, this is not because of our method. % our methods fault. Because,
As discussed in Section \ref{subsubsec:imageimage} the CycleGAN method could not provide the same results as the previous 
work and even with a $3.3\%$ improvement with our method, the result still is bellow expected.

In summary our method is significantly better than direct transfer without adaptation.
It is important to emphasize that our method does not make use of any label from the target domain, completely removing the burden of annotating new data when the application domain changes.

\subsection{Batch scheduler results}
\label{subsec:result_batch}

In order to analyze the batch scheduler contribution, we performed experiments with and without the batch scheduler algorithm
using the new method (based on AlignedReID++) and domain adaptation with CycleGAN and CycleGAN\&pseudo-labels (Ours).
We have not performed experiments with the batch scheduler for direct transfer because for the Market1501 and CUHK03 datasets we used pre-trained weights from the AlignedReID++ paper \cite{luo2019alignedreid++} and the Viper dataset does not have enough data to profit from the batch scheduler algorithm.

Considering only the rank-1 results shown in Table \ref{table:results}, 
we have 8 test cases where it was better not to use the batch scheduler and 4 test cases that indicate the opposite.
Although the majority of test cases indicates that the batch scheduler does not help, 4 of these 8 cases use the
Viper images for training (adapted or not).
The problem is that the Viper dataset has only 1264 images, then the assumption that we made in Eq.~\ref{eq:noise} when we said
that $N$ was big enough to approximate the Equation to $g \approx \varepsilon N/B$ does not hold for this dataset.
Because of that, the $-1$ factor in Eq.~\ref{eq:noise} has a strong contribution and the assumption $g \propto 1/B$ is not valid, but
$g \propto \varepsilon$ is correct. Therefore, a classical learning rate decay scheduler works better in these cases.

Having this limitation of the Viper dataset in mind, we can focus our analysis on the experiments that
did not involve that dataset. However, that still gives a draw of 4 cases in favour and 4 cases against the
batch scheduler.

Our results are therefore inconslusive regarding the batch scheduler. We hypothesise that a major factor for that is that we used a GPU with an amount of memory that was too small (8GB) to be effective for this strategy, allowing a maximum batch size of 88 samples.
It remains as future work to evaluate this on better hardware or with a memory management strategy that would allow larger batches with limited hardware.

\section{Conclusions}
\label{sec:conclusion}

 In person re-identification, each type of environment (e.g.\ airport, shopping center, university campus, etc.)
has its own typical appearance, so a system that is trained in one environment may not perform well in another environment.
This observation was confirmed by our cross-dataset (direct transfer) experiments, indicating that each dataset can be treated as a domain.
We showed that a domain adaptation method based on CycleGAN can be applied to transform the marginal distribution of samples from
a source dataset to a target dataset. This enables us to retrain a triplet CNN on adapted samples so that their performance
is improved on the target dataset without using a single labeled sample from the target set.
Furthermore, we showed that using this CNN and a clustering algorithm to generate pseudo-labels and retrain the triplet CNN leads to a significant boost in the performance on target dataset.
This opens doors for the deployment of person re-ID software to real applications, as it completely removes the burden of annotating new data.

Further to proposing a domain adaptation technique for this problem, we also presented the use of a batch scheduler which increases the batch size as training starts to converge.
However, the hardware limitations and the lack of data in Viper dataset prevented us from a deep analysis of this method's effectiveness.

In addition, this paper proved that our method can be applied with state-of-art person re-identification methods as backbone (AlignedReID++). Also, it was clear that
the better the backbone method, the better are the results achieved with our workflow.

For future work, we believe it would be interesting to try our technique with other datasets and investigate other ways to group samples and create pseudo-labels.
% But it is proved that this technique brings great contribution to the field of person re-identification.

\section*{Acknowledgements}

 The authors would like to thank FAPDF (\href{http://www.fap.df.gov.br}{fap.df.gov.br}) and CNPq grant PQ 314154/2018-3 (\href{http://cnpq.br/}{cnpq.br}).

\newpage

\bibliography{defs,bibliography}
\bibliographystyle{IEEEtran}

\end{document}